\RequirePackage{fix-cm}
\documentclass[twocolumn]{svjour3}          
\smartqed  
\usepackage{graphicx}
\usepackage{hyperref}
    \hypersetup{
        colorlinks   = true,
        citecolor    = blue,
        linkcolor=blue  
    }

\usepackage{xstring}
\makeatletter
\AtBeginDocument{
\let\oldref\ref
\renewcommand{\ref}[1]{\IfBeginWith{#1}{fig:}%
{{\color{blue}Fig.~\oldref{#1}}}%
{\IfBeginWith{#1}{tab:}{{\color{blue}Table~\oldref{#1}}}%
{\IfBeginWith{#1}{sec:}{{\color{blue}Section~\oldref{#1}}}%
{\IfBeginWith{#1}{Eq:}{{\color{blue}Eq.~\oldref{#1}}}%
{\IfBeginWith{#1}{appendix1}{{\color{blue}Appendix 1}}}%
{}}}
}}}
\makeatother

\usepackage{caption}
\usepackage{subcaption}
\usepackage{siunitx}
\usepackage{float}
\usepackage{comment}
\usepackage{balance}
\usepackage[dvipsnames]{xcolor}

\begin{document}

\title{A Skip-connected Multi-column Network for Isolated Handwritten Bangla Character and Digit recognition}


\author{Animesh Singh \and Ritesh Sarkhel \and Nibaran Das \and Mahantapas Kundu \and Mita Nasipuri
}
\institute{Animesh Singh \at
             Department of Computer Science and Engineering \\
             Jalpaiguri Government Engineering College, West Bengal \\
              \email{as2002@cse.jgec.ac.in}           
           \and
            Ritesh Sarkhel \at
            Department of Computer Science and Engineering \\
            The Ohio State University, Colombus, Ohio \\
            \email{sarkhel.5@osu.edu}
            \and
            Nibaran Das \at
            Department of Computer Science and Engineering \\
            Jadavpur University, Kolkata, West Bengal \\
            Phone no: +913324572407 \\
            \email{nibaran.das@jadavpuruniversity.in}
            \and
            Mahantapas Kundu \at
            Department of Computer Science and Engineering \\
            Jadavpur University, Kolkata, West Bengal \\
            \email{mahantapas.kundu@ jadavpuruniversity.in}
            \and
            Mita Nasipuri \at
            Department of Computer Science and Engineering \\
            Jadavpur University, Kolkata, West Bengal \\
            \email{mita.nasipuri@jadavpuruniversity.in}
}
\date{Received: date / Accepted: date}
\maketitle

\begin{abstract}
Finding local invariant patterns in handwritten characters and/or digits for optical character recognition is a difficult task. Variations in writing styles from one person to another make this task challenging.We have proposed a non-explicit feature extraction method using a multi-scale multi-column skip convolutional neural network in this work. Local and global features extracted from different layers of the proposed architecture are combined to derive the final feature descriptor encoding a character or digit image. Our method is evaluated on four publicly available datasets of isolated handwritten Bangla characters and digits. Exhaustive comparative analysis against contemporary methods establish the efficacy of our proposed approach.

\end{abstract}

\section{Introduction}

Optical character recognition (OCR) denotes the process of automatically recognizing characters from an optically scanned page of handwritten or printed texts. Some popular real world applications of OCR includes document information extraction \cite{sarkhel2019visual,wu2018fonduer}, question answering \cite{andreas2016learning,dong2015question}, classification of documents as potential applications \cite{sarkhel2019deterministic,das2018document} etc. Recognition of handwritten characters is a difficult task as finding the local invariant patterns within handwritten character or digit images is not trivial. The already difficult task of OCR gets more challenging in case of stylistically varying handwritten texts as handwriting style varies from one person to another. In automated OCR systems, features describing the local invariant patterns are extracted and sophisticated classification models are employed to recognize a handwritten character or digit. Such feature-based OCR systems can be classified into two major categories: \textit{explicit feature-based} and \textit{non-explicit feature-based} OCR systems. Syntactic or formal grammar-based features \cite{feng1975decomposition}, moment-based features \cite{singh2016study}, graph-theoretic approaches \cite{kahan1987recognition}, shadow-based features \cite{basu2009hierarchical}, gradient-based features \cite{roy2012new} etc. are some of the most popular examples of explicit feature-based OCR system. In this approach handcrafted features are explicitly designed and extracted from a character or digit image. Contrary to the explicit feature-based approaches, non-explicit feature (\textit{NeF}) based methods do not rely on handcrafted features. {As raw character or digit images are fed into the OCR, a learned model automatically extracts features from the image. The model itself is learned over multiple iterations by minimizing the misclassification error.} Artificial neural network \cite{le1994word,basu2012mlp}, deep convolutional neural network based approaches \cite{pratt2019handwritten,chowdhury2019bangla,ukil2019improved}, recurrent neural network \cite{ren2019recognizing,paul2019recognition,ahmed2019handwritten}, markov-model based approaches \cite{britto2003recognition,bunke2004offline} and unsupervised sparse feature extraction \cite{hasasneh2019towards} methods are some examples of non-explicit feature (\textit{NeF}) based OCR systems. Statistically derived parameters play a significant role in those approaches.


\begin{figure*}
\centering \makeatletter\IfFileExists{HWCAD.jpg}{\includegraphics[width=0.9\textwidth]{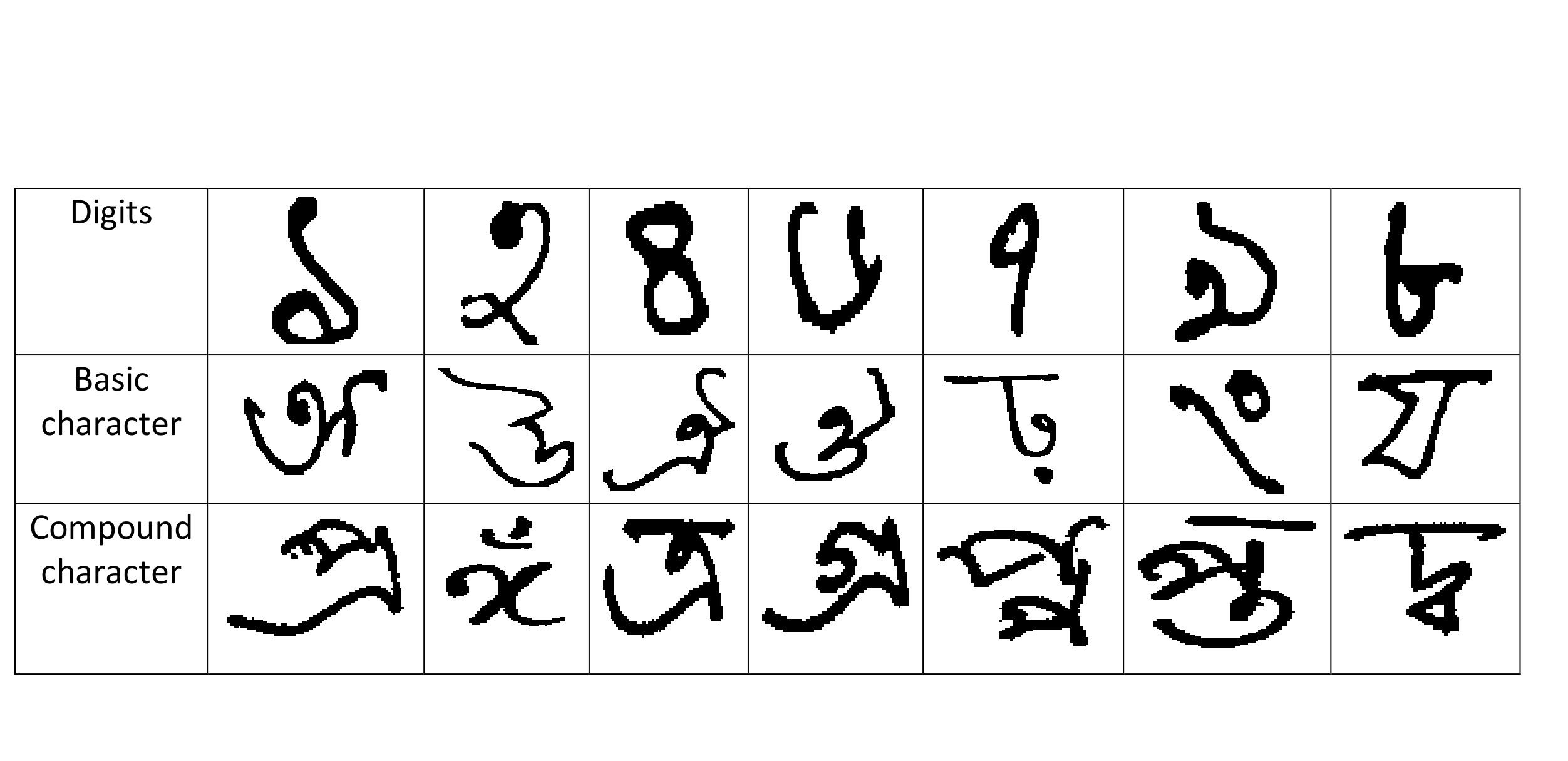}}
\makeatother 
\caption{A few samples of Bangla handwritten characters and digits}
\label{fig:HWCAD}
\end{figure*}

Based on spatial coverage two types of features have traditionally been used in explicit and non-explicit feature based OCR systems, local features and global features. Local features refer to a pattern or distinct structure found in a small image patch. They are usually associated with local properties of an image patch that differs from its immediate surroundings in texture, colour, or intensity. Examples of local features used in contemporary literature include blobs, corners, and edge pixels. Local features have been used in a wide variety of applications. A \textit{sampling} based approach \cite{nowak2006sampling}, proposed by Nowak et al., extracts visual descriptors for classification task by sampling the independent image patches. Haralick et al. has proposed a \textit{textural-feature} based approach \cite{haralick1973textural} which uses a piecewise liner decision rule and a min-max decision rule for identification of image data. A \textit{multi-scale differential} model \cite{schmid1997local}, developed by Schmid et al., uses a voting algorithm with semi-local constraints for retrieving images from large image databases. A \textit{deep convolutional feature} based approach \cite{babenko2015aggregating}, proposed by Babenko et al., extracts local deep features from images with the help of a deep convolutional neural network. These local features are finally aggregated to produce compact global descriptors for image retrieval task. Hiremath et al. has proposed a \textit{contend based} approach \cite{hiremath2007content} for retrieval of images. This approach uses primitive local image descriptors like color moments, texture and shape for image retrieval. A \textit{hierarchical agglomerative clustering} based model \cite{mikolajczyk2005local}, developed by Mikolajczyk et al., combines local features of different objects using an entropy distribution for recognizing multiple object classes. An optimization approach \cite{jiang2007towards}, proposed by Jiang et al. tries to find the optimal combination of detector, kernel, vocabulary size and weighting scheme for \textit{bag-of-features (BoF)}. These aforementioned works are some popular examples of usage of local image features.

Global features, on the other hand, describe the entire image as a whole, representing the shapes and contours present in the image. Global features have also been used widely in the area of computer vision. Wang et al. has proposed a \textit{combination technique} \cite{wang2009combining} which extracts global, regional and contextual features from images and accumulates these extracted features by estimating their joint probability for automatic image annotation. Shyu et al. has developed a \textit{content-based} image retrieval model \cite{shyu1998local} which uses low-level computer vision technique and image processing algorithms for extracting features related to the different variations of gray scale, texture, shape etc. Labit et al. proposed a \textit{compact motion representation based} method for semantic image sequence coding \cite{labit1991compact} which uses global image features. These are some popular examples of usage of local image features.

Both local and global image features play important roles in automated handwritten character and digit recognition systems. {A region sampling based approach \cite{das2012genetic,sarkhel2015enhanced} to extract local features from the most discriminating regions of a character/digit image, an ensemble technique \cite{das2012statistical} to combine quad-tree based longest-run features with statistical features using \textit{PCA} \cite{wold1987principal} and \textit{Modular PCA} \cite{sankaran2004multi}, artificial neural network-based approach \cite{das2010handwritten,basu2012handwritten,basu2012mlp}, deep convolutional neural network-based approach \cite{benaddy2019handwritten,roy2017handwritten}, multilayered boltzmann perceptron network \cite{rehman2019cursive} to extract hard geometric features, Transfer learning based approach \cite{chatterjee2019bengali} which used a pre-trained ResNet \cite{he2016deep}, a \textit{multi-column} based approach \cite{sarkhel2017multi} to extract multi-scale local features from pattern images for recognition of handwritten characters and digits, a multi-layer capsule network based approach \cite{mandal2019handwritten} etc are some examples of local feature extraction based handwritten character/digit recognition methods. Global image features have also been used for this purpose such as \textit{convex hull-based} features \cite{sarkhel2016multi,das2014recognition} and \textit{chain code histogram-based} features \cite{bhattacharya2006recognition}, artificial neural network based approach \cite{singh2014handwritten} etc. In some these aforementioned works a combination of local and global image features have also been used but in most of the works global features are extracted through an explicit based feature extraction technique. Designing a non-explicit based both global and local feature extraction method for recognition of handwritten character or digit is the primary concern of our present work.}

\begin{figure*}[t]
\centering \makeatletter\IfFileExists{CNN_architecture.JPG}{\includegraphics[width=\textwidth]{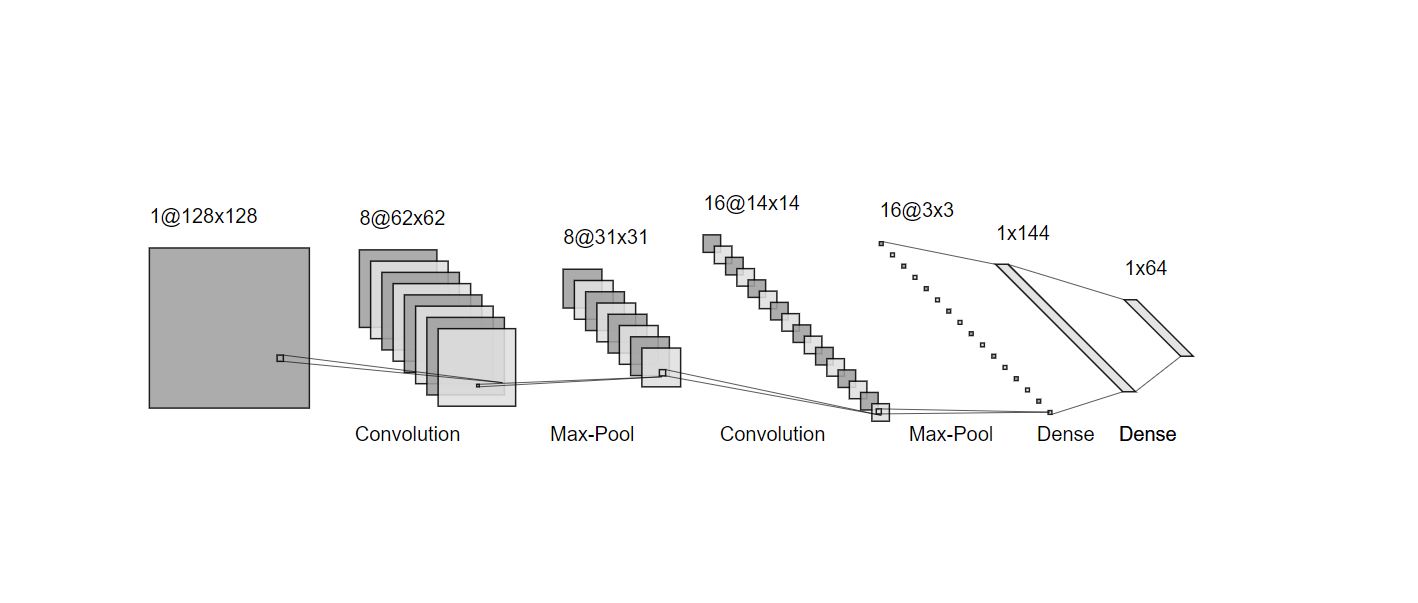}}{}
\makeatother 
\caption{{A traditional convolutional architecture which consists of several \textit{convolutional} and \textit{max-pooling} layers stacked on the top of each other}}
\label{fig:CNN_architecture}
\end{figure*}

Despite the progress on developing OCR systems for Indic scripts in the past decade, a commercially successful comprehensive system is yet to emerge. In this work, we have proposed an \textit{NeF} based approach for recognizing handwritten Bangla digits and characters (shown in \ref{fig:HWCAD}) by combining global and local features extracted using a deep convolutional network (CNN). We have developed a multi-column skip-connected convolutional neural network (MSCNN) architecture for this purpose. Global features extracted from the initial layers of the network are combined with local features from the final layers of this convolutional architecture. These features are learned by training the network over multiple iterations to minimize misclassification error. We proposed a novel fusion technique to combine these global and local features extracted from different layers of the architecture to generate the final feature descriptor representing a character or a digit image. We have evaluated the proposed method on five publicly available, benchmark datasets of handwritten Bangla characters and digits. Promising results have been achieved for all the datasets. We have also tested our system on MNIST dataset and achieved a maximum accuracy of 99.65~\%, without any data augmentation to the original data. A comparative analysis has also been performed against some of the contemporary methods to establish the superiority of our proposed method.
The rest of the paper is organized as follows: \ref{sec:A Brief Overview Of CNN} introduces a brief overview of CNN based architectures, \ref{sec:Present Work} presents a detailed description of the present work, the datasets used in the experimental setup and the experimental results are described in \ref{sec:Experimental Results}. A comparative study based on different possible combination technique is presented in \ref{sec:Comparative Analysis} and finally, a brief conclusion is made from the results.

\section{A brief overview of CNN}\label{sec:A Brief Overview Of CNN}
\begin{figure}[t]
\centering \makeatletter\IfFileExists{Working_of_CNN.JPG}{\includegraphics[width=1\linewidth]{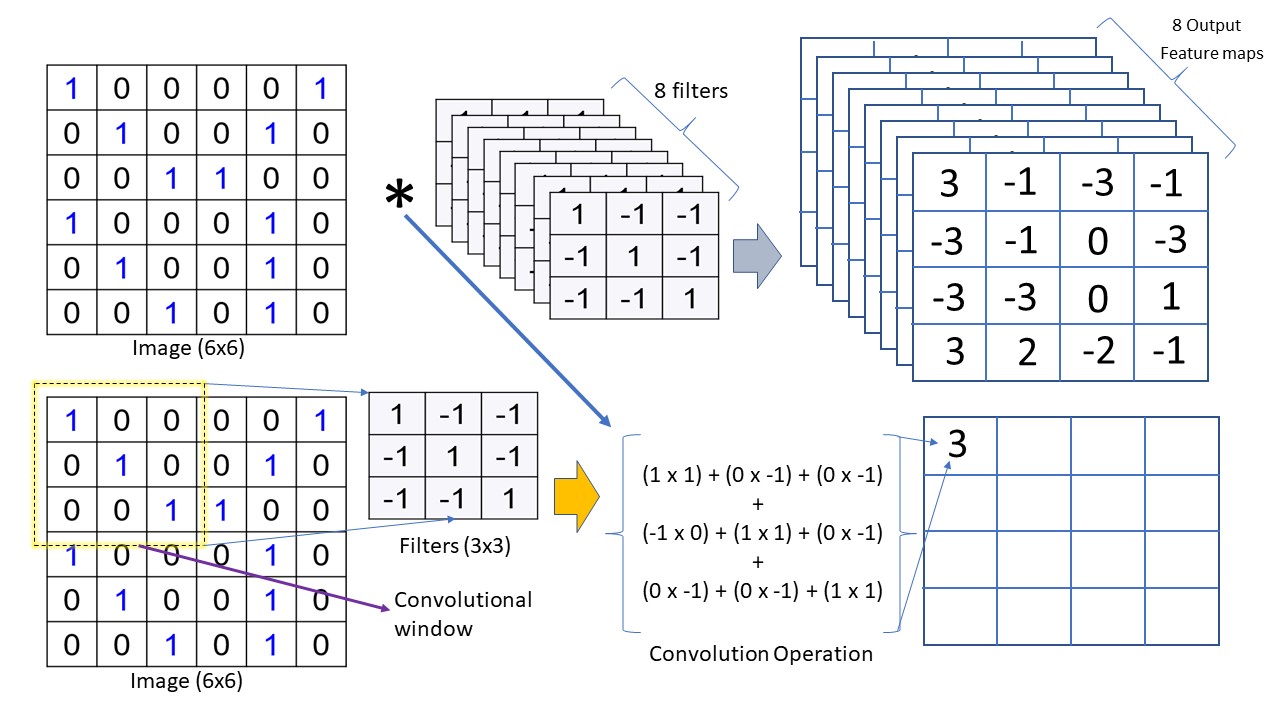}}{}
\makeatother 
\caption{{The \textit{convolutional operation} performed by CNN layers}}
\label{fig:Working_of_CNN}
\end{figure}

Convolutional neural networks (CNNs) \cite{le1994word,roy2017handwritten} are a class of feedforward networks generally used for recognizing patterns of an image. Just as artificial neural network (ANN), CNNs are also biologically inspired architectures. The hierarchical information processing style by the alternating layers of simple and complex cells of visual cortex in brain \cite{serre2006object} motivates the architecture of CNNs. In general, the architectures consist of several \textit{convolutional }and \textit{pooling }(or subsampling) layers stacked on the top of each other as shown in \ref{fig:CNN_architecture}. The convolutional layers learn the features representing the structures of the input image and thus, they serve as feature extractors. Each neuron in the convolutional layers relates to the neighborhood neurons of the previous layers via a set of learnable weights; also known as filter banks (LeCun et al. \cite{lecun2012efficient}). These neurons in the convolutional layers form feature maps by arranging in a specific order. Patterns in input images are extracted with the learned weights in order to compute a new feature map (shown in \ref{fig:Working_of_CNN}) and this map is forward propagated into a non-linear activation function, which allow extracting non-linear features. {Different parts of a feature map share similar weights to learn translational invariant features and different feature maps from same convolutional layer contain different weights which help learning multiple patterns from every part of a pattern image.} Pooling layers play a significant role to achieve spatial invariance to input distortions and translations. Initially, most common pooling operation was performed with average pooling \cite{boureau2010theoretical}, in which the average of all the inputs values over a small region of an image is propagated to the next layer. However, the most recent works {(Tolias et al. \cite{tolias2015particular}; Giusti et al. \cite{giusti2013fast}; Nagi et al. \cite{nagi2011max}; Scherer et al. \cite{scherer2010evaluation}; Murray et al. \cite{murray2014generalized} etc)} generally used max pooling \cite{nagi2011max,giusti2013fast}, in which maximum of all the input values is propagated to the next layer. Hence, convolutional layers and pooling layers are two main building blocks of CNN architectures. Therefore, the local invariant patterns are collected by the convolutional layers and processed by pooling layers to extract more powerful features.

\begin{figure}
\centering \makeatletter\IfFileExists{Multi_Column.jpg}{\includegraphics[width=1\linewidth]{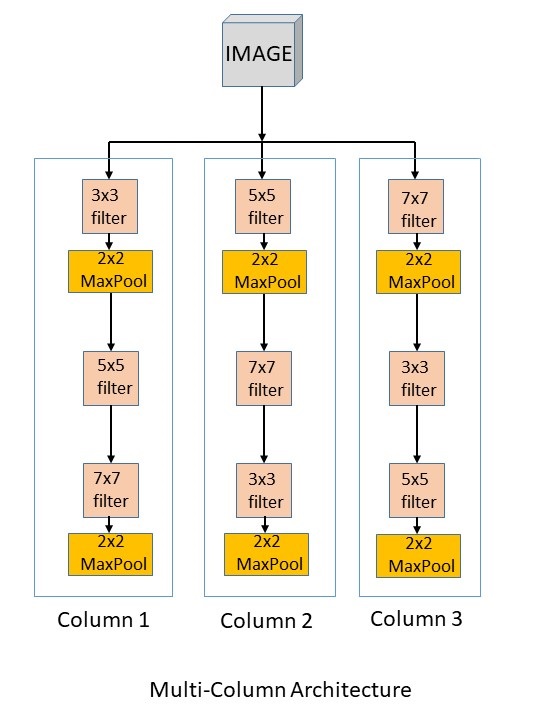}}{}
\makeatother 
\caption{{Multi-column based CNN architecture in which \textit{level-wise multi-scaling} and \textit{column-wise multi-scaling} is used}}
\label{fig:Multi_Column}
\end{figure}

In the field of OCR system, researchers \cite{sarkhel2017multi} have found in their experiments that convolutional sampling at fixed scales often bounds the local invariant pattern searching capabilities of a CNN; but sampling at multiple scales allows a CNN to extract more robust and noise invariant features from different patches of an image. In order to address this scope, the filter banks of the proposed architecture have variable sizes (i.e. $3\times3$, $5\times5$, $7\times7$) at different convolutional layers (as shown in \ref{fig:Multi_Column}). As, the maximum information loss occurs due to the pooling layers, we have used smaller window sizes (i.e. $2\times2$) in our proposed system.

\begin{figure*}
\centering \makeatletter\IfFileExists{combination.jpg}{\includegraphics[width=0.9\textwidth]{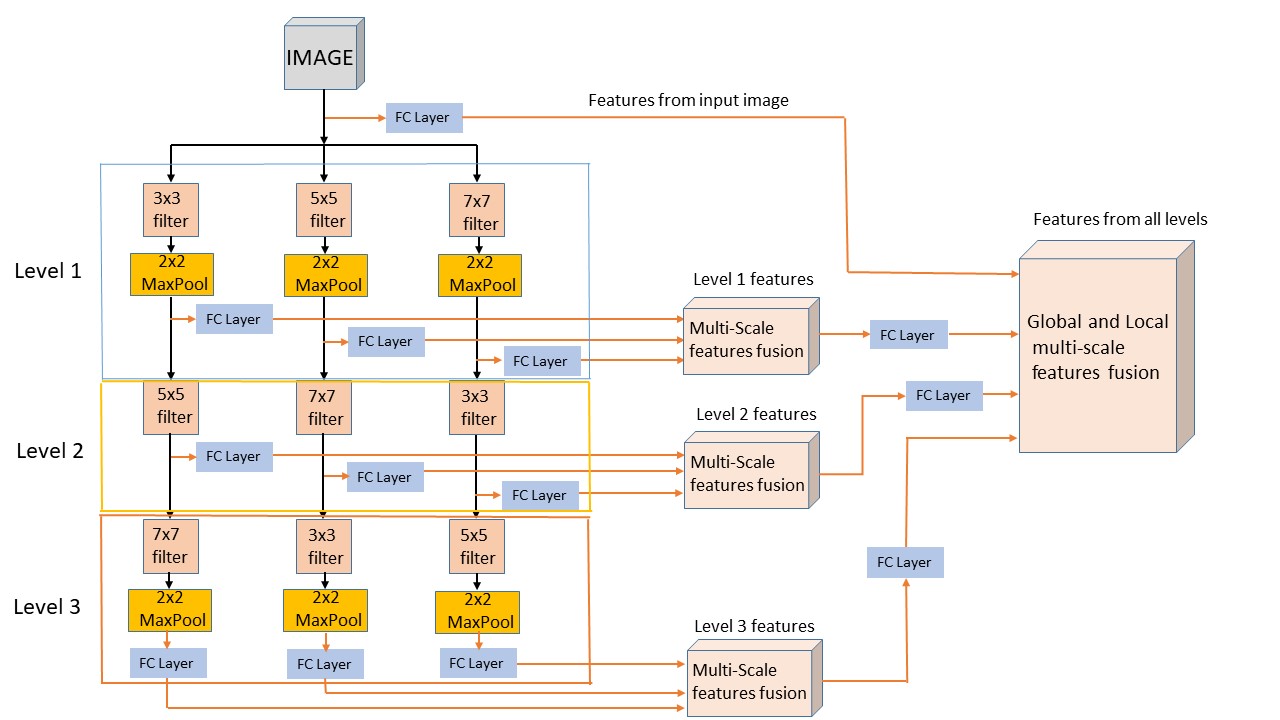}}{}
\makeatother 
\caption{{A \textit{multi-column} based architecture used in our proposed system}}
\label{fig:combination}
\end{figure*}

Multi-column convolutional neural network (MCNN) (inspired by the microcolumns of neurons in the cerebral cortex) based architecture has shown a major breakthrough in various research works like handwritten text recognition \cite{basu2012handwritten,singh2014handwritten,cirecsan2012multi,sarkhel2017multi}, traffic sign classification \cite{ciresan2011committee}, crowd counting \cite{zhang2016single,sam2017switching} etc. Multiple columns of feedforward convolutional neural networks are used together to combine the output feature maps from all the columns. Different feature maps are generated for same image patches at different columns because the weights of the filter banks in convolutional layers vary from one column to other. When a pattern image is propagated through every column of a MSCNN based architecture, multiple feature maps from each image patch are extracted at the convolutional layers of the respective columns. Feature maps generated at one convolutional layer of a definite column differs from the similar (layers at same depth) convolutional layers of other columns. Similarly, for other layers of different columns, several variations of feature maps are created and when these multi-variate feature maps are combined more robust abstraction of a pattern image is created and provides a better prediction accuracy in a OCR system.

In multi-column CNN based architecture, multi-scaling can be applied in two different ways, one is \textit{column-wise multi-scaling }and another is \textit{level-wise multi-scaling.} In \textit{column-wise multi-scaling, }multiple scales of convolutional filters (i.e. variable filter sizes used in convolutional layers) are used within each column of MSCNN based architecture, and more robust and geometrically invariant patterns are extracted from different patches of images within a single column \cite{sarkhel2017multi}. {On the other hand, in \textit{level-wise multi-scaling} similar types {(layers at same depth)} of convolutional layers of different column has different filter sizes (shown in \ref{fig:Multi_Column}), hence multi-variate patterns are sampled at the same levels of different columns. The patterns derived from these two types of multi-scaling techniques when accumulated create shift, scale and distortion invariant feature maps that contain more complex and deeper geometrical information of character or digit images.}

\section{Present work}\label{sec:Present Work}
The primary contribution of our present work is to utilize the information contained in both global and local features of pattern images in handwritten characters and digits recognition tasks. As mentioned before, global features define a geometrical abstraction of the entire image and local features define more detailed (upto pixel level) description of an image. Hence, global features are more sensitive towards clutter and occlusion, but local features are more robust in those cases \cite{mikolajczyk2005local}. In our proposed system the global and the local features are extracted in a non-explicit based feature extraction approach from different parts of our network. From our experiments we observed that, the connection weights between the anterior layers of the network are more prone to learn global features and the connection weights between the posterior layers of the network are more likely to extract local features from pattern images. {These global and local features, when gathered together, contain more robust and accurate description of pattern images  and significantly improves recognition results over the methods which use either global or local features alone.} A graphical representation of how global and local features are fused in our present work is shown in \ref{fig:combination}.

As mentioned above a \textit{NeF} based approach using MSCNN based architecture for the recognition of isolated handwritten characters and digits of popular Bangla scripts has been proposed in the present work. Combination of features extracted at final FC layer of every column of the multi-column based architecture is used as implicit feature descriptor of pattern images. After learning the connection weights between the layers through training; the test image is forward propagated through the network and the final feature descriptor are extracted between the final fully-connected (FC) layer and the softmax classifier. Finally, the softmax classifier performs a classification task using the final extracted feature to predict the final class label. A graphical abstract of the proposed architecture is shown in \ref{fig:Architecture}. More detailed description is presented in \ref{sec:Description Of Proposed Architecture} and \ref{sec:Architecture Of Each Column}. As mentioned in the previous section, the proposed system uses two different types of multi-scaling techniques: \textit{column-wise multi-scaling} and \textit{level-wise multi-scaling}, which enhance the global and local invariant pattern searching capabilities of the proposed architecture from character or digit images. {When these multi-scaled local and global features are combined improves the recognition performance of our proposed network on multiple intricate handwritten character and digit images.}

\subsection{Description of the proposed architecture}\label{sec:Description Of Proposed Architecture}
\begin{figure}
\centering \IfFileExists{concatenation.jpg}{\includegraphics[width=\linewidth]{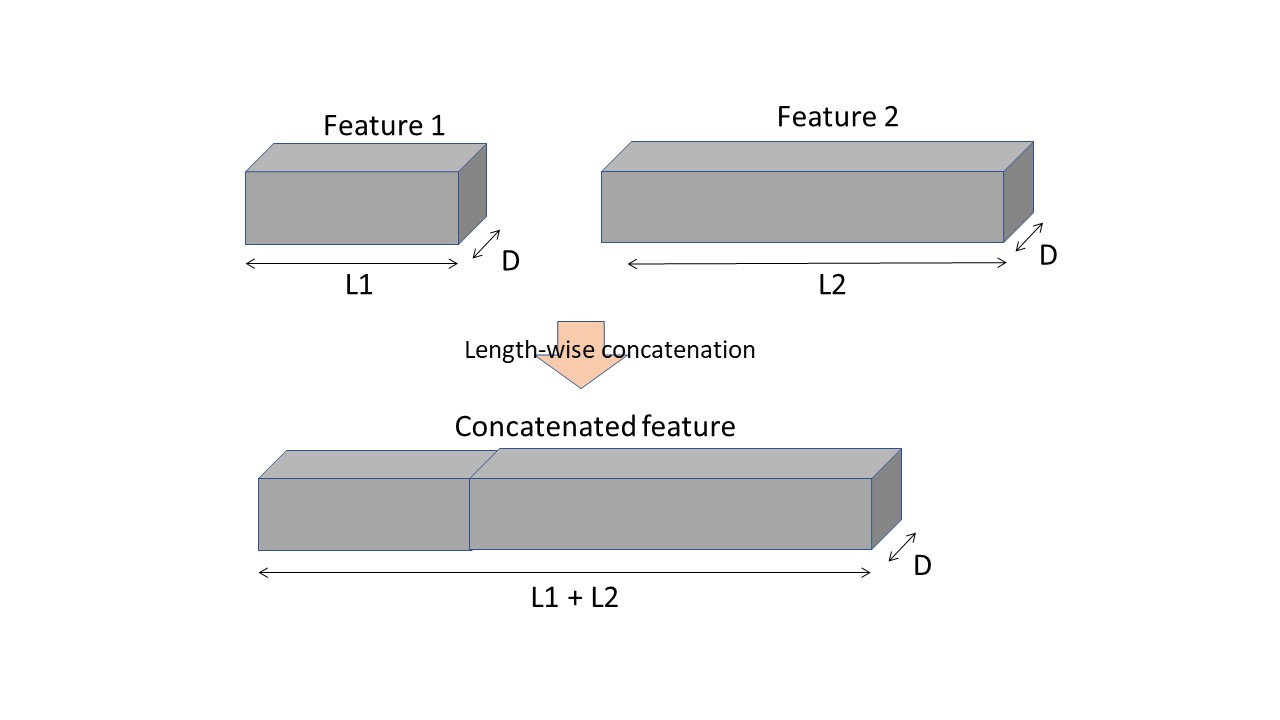}}{}
\makeatother 
\caption{{\textit{Feature concatenation} method used in our present work to combine different types of features coming from different parts of the present architecture}}
\label{fig:concatenation}
\end{figure}

\begin{figure*}
\centering \makeatletter\IfFileExists{Architecture.jpg}{\includegraphics[width=1\textwidth]{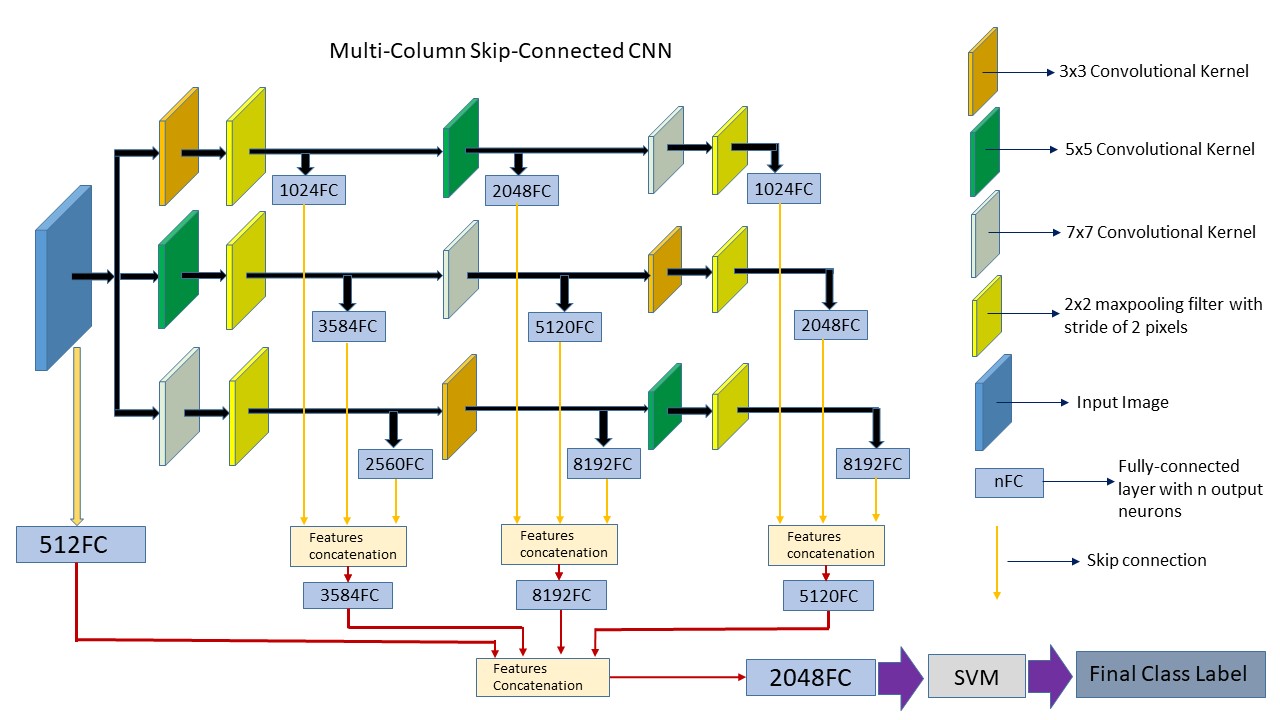}}{}
\makeatother 
\caption{{A graphical abstract of the \textit{proposed architecture}. The architecture uses \textit{global} and \textit{local }features extracted by a \textit{multi-column multi-scale convolutional neural network}}}
\label{fig:Architecture}
\end{figure*}

Details of the MSCNN-based architecture is presented in this section. The architecture contains three columns with a subtle architectural difference among them. Architecture of each of the columns has been configured empirically. Each column has three levels and each of the levels consists of a single convolutional layer or a stack of convolutional and pooling layers as shown in \ref{fig:combination}. From each level of every column, the output feature map is forward propagated through a fully-connected (FC) layer to extract more useful and necessary features from the feature maps extracted at convolutional layers (shown in \ref{fig:Architecture}). Without these FC layers, if the feature maps from the convolutional layers are directly used, the network may learn redundant and confusing features for updating the connection weights. Hence, the FC layers play a significant role to eradicate some of those unnecessary features from the feature maps extracted at different levels of the network. The configurations of these local FC layers are also finalized empirically. {The feature generation process is explained in \ref{Eq:11} - \ref{Eq:F33} and feature combination process is described in \ref{Eq:W1} - \ref{Eq:FF}}. The proposed methodology combines multi-scaled global and local invariant features sampled at multiple layers of the network through a \textit{features-concatenation} method (shown in \ref{fig:concatenation}).

\begin{small}
\begin{equation}
    X_{11} = C_{11}(I) \label{Eq:11}
\end{equation}

\begin{equation}
    X_{12} = C_{12}(I) \label{Eq:12}
\end{equation}

\begin{equation}
    X_{13} = C_{13}(I) \label{Eq:13}
\end{equation}

\begin{equation}
    Y_{11} = F_{11}(X_{11}) \label{Eq:F11}
\end{equation}

\begin{equation}
    Y_{12} = F_{12}(X_{12}) \label{Eq:F12}
\end{equation}

\begin{equation}
    Y_{13} = F_{13}(X_{13}) \label{Eq:F13}
\end{equation}

\begin{equation}
    X_{21} = C_{21}(C_{11}(I)) \label{Eq:21}
\end{equation}

\begin{equation}
    X_{22} = C_{22}(C_{12}(I)) \label{Eq:22}
\end{equation}

\begin{equation}
    X_{23} = C_{23}(C_{13}(I)) \label{Eq:23}
\end{equation}

\begin{equation}
    Y_{21} = F_{21}(X_{21}) \label{Eq:F21}
\end{equation}

\begin{equation}
    Y_{22} = F_{22}(X_{22}) \label{Eq:F22}
\end{equation}

\begin{equation}
    Y_{23} = F_{23}(X_{23}) \label{Eq:F23}
\end{equation}

\begin{equation}
    X_{31} = C_{31}(C_{21}(C_{11}(I))) \label{Eq:31}
\end{equation}

\begin{equation}
    X_{32} = C_{32}(C_{22}(C_{12}(I))) \label{Eq:32}
\end{equation}

\begin{equation}
    X_{33} = C_{33}(C_{23}(C_{13}(I))) \label{Eq:33}
\end{equation}

\begin{equation}
    Y_{31} = F_{31}(X_{31}) \label{Eq:F31}
\end{equation}

\begin{equation}
    Y_{32} = F_{32}(X_{32}) \label{Eq:F32}
\end{equation}

\begin{equation}
    Y_{33} = F_{33}(X_{33}) \label{Eq:F33}
\end{equation}
\end{small}

Some notations are used to describe the procedure of feature combination after each level of the proposed MSCNN based architecture. $C_{ij}(X)$ represents the the output function of convolutional layer (along with maxpool layer) at level $i$ and column $j$ of the proposed architecture, where $C_{ij}$ denotes the corresponding convolutional and maxpooling layer and $X$ denotes the input. $F_{mn}(Z)$ represents the output function of fully-connected layer at level $m$ and column $n$ of the proposed architecture, where $F_{mn}$ denotes corresponding FC layer with input $Z$, where $Z$ is $C_{mn}(X)$. {$F_{k}(W)$ represents the output function of final fully-connected layer at column level $k$ with input $W$, where $W$ is the concatenated features from all the columns of level $k$. $F(I)$ represents the output function of the fully-connected layer where the input is original image $I$. $Q_{f}(G)$ represents the output function of final FC layer of the proposed architecture, where $G$ is the final concatenated feature from different levels of the network. Feature concatenation is denoted by $\oplus$.} Input image is forward propagated through every column and different feature maps are generated at different levels of the architecture. {$C_{11}$, $C_{12}$, $C_{13}$ extract features from the input images and the local FC layers $F_{11}$, $F_{12}$, $F_{13}$ extract more useful features from the feature maps extracted at the previous sampling layers. Similarly, $F_{21}$, $F_{22}$, $F_{23}$ extract features from the feature maps sampled at $C_{21}$, $C_{22}$, $C_{23}$ in the second column and $F_{31}$, $F_{32}$, $F_{33}$ extract features from the feature maps sampled at $C_{31}$, $C_{32}$, $C_{33}$ in the third column. The output features of each local FC layer at same levels of different columns are combined by the proposed \textit{features-concatenation} technique (shown in \ref{fig:combination}). This means features from first levels of all the columns (i.e. output features of $F_{11}$, $F_{12}$, $F_{13}$) are combined, similarly features from the second levels of all the columns (i.e. output features of $F_{21}$, $F_{22}$, $F_{23}$) are combined and features from the third levels of all the columns (i.e. output features of $F_{31}$, $F_{32}$, $F_{33}$) are combined. These concatenated features at different levels are propagated through $F_{1}$, $F_{2}$, $F_{3}$ layers, and finally, the output features of $F_{0}$, $F_{1}$, $F_{2}$, $F_{3}$ layers are combined with feature concatenation method to generate the final feature descriptor($Q_{f}(G)$).} With this feature combination method, basically, we are trying to assemble similar kind of features i.e. features coming from equal numbers of sampling (convolutional and pooling) layers. This methodology helps to accommodate several variations of patterns from different image patches into a single bank of feature descriptors. These feature banks created at different levels contain features sampled at multiple scales, hence provide a better geometrical abstraction of pattern images. At the first level the features are extracted from one convolutional layer, similarly at the second level the features are extracted from two convolutional layers and one maxpooling layer and at the final level the features are extracted from three convolutional layers and two maxpooling layers. All the features from the same levels do not represent the exact same abstraction of the actual image because different columns have different filter sizes in their respective convolutional layers. Combination of these similar kind of features denotes one of the significant contribution of the present work.\pagebreak

\begin{small}
\begin{equation}
    W_{1} = Y_{11} \oplus Y_{12} \oplus Y_{13} \label{Eq:W1}
\end{equation}

\begin{equation}
    W_{2} = Y_{21} \oplus Y_{22} \oplus Y_{23} \label{Eq:W2}
\end{equation}

\begin{equation}
    W_{3} = Y_{31} \oplus Y_{32} \oplus Y_{33} \vspace{0.2cm} \label{Eq:W3}
\end{equation}

\begin{equation}
    G = F(I) \oplus F_{1}(W_{1}) \oplus F_{2}(W_{2}) \oplus F_{3}(W_{3}) \label{Eq:FF}
\end{equation}
\end{small}

\subsection{Architecture of each column}\label{sec:Architecture Of Each Column}\vspace{-0.2cm}
The configuration of every column is given in this section. The first column has the following architecture: 32C2-2P2-BN-RELU-64C1-BN-RELU-256C1-2P2-BN-\\RELU. In this representation XCY denotes a convolutional layer with a total of X kernels and stride of Y pixels, MPN denotes a max-pooling layer with an $M \times M$ pooling window and stride of N pixels, BN denotes Batch-Normalization \cite{ioffe2015batch}, RELU denotes a Rectified Linear Unit activation layer \cite{maas2013rectifier,dahl2013improving}. Every level of the first column has a definite FC layer. In the first column, after the first level 1024FC-BN-RELU layer is used; similarly, after the second level and the third level 2048FC-BN-RELU layer and 1024FC-BN-RELU layer are used respectively. In this representation NFC denotes a fully-connected layer with N numbers of output neurons. The second column has the following architecture: 32C1-2P2-BN-RELU-64C1-BN-RELU-256C2-2P2-BN-RELU. In the second column, after the first level, second level and third level 3584FC-BN-RELU layer, 5120FC-BN-RELU layer and 2048FC-BN-RELU layer are used respectively. And the third column has the following architecture: 32C1-2P2-BN-RELU-64C1-BN-RELU-256C1-2P2-BN-RELU. In the third column, after the first level, second level and third level 2560FC-BN-RELU layer, 8192FC-BN-RELU layer and 8192FC-BN-RELU layer are used respectively. A graphical representation of the architecture is shown in \ref{fig:Architecture}.

From first level of all the columns, the extracted features from the local FC layers are combined and the concatenated features are forward propagated through 3584FC-BN-RELU layer. Similarly, from the second and third level the extracted features are combined and forward propagated through 8192FC-BN-RELU layer and 5120FC-BN-RELU layer respectively. And the actual image is also directly passed through a 512FC-BN-RELU layer. Output features from these four outer fully-connected layers are again combined through features-concatenation method (shown in \ref{fig:concatenation}) and finally the resultant feature vector is propagated through 2048FC-BN-RELU-softmax layer. An RBF kernel based multi-class SVM (One vs All) classifier \cite{niu2012novel} is used as softmax classifier in the present architecture and by performing $\beta$/$\gamma$ variations the parameters of the kernel are tuned empirically. As mentioned before, three different convolutional kernel dimensions i.e. $3 \times 3$, $5 \times 5$, $7 \times 7$ (as in \ref{fig:Multi_Column}) are used in our present work for multi-scaling feature extraction.\vspace{-0.1cm}

\subsection{Training the proposed architecture}
The proposed architecture is trained with an end-to-end training method in which all the three columns is trained simultaneously. The training images are passed through all the columns simultaneously and the loss is calculated at the final FC layer. {The connection weights between the layers of the whole architecture are updated in a single pass of backpropagation after every epoch using equations \ref{Eq:Rmsprop1}, \ref{Eq:Rmsprop2}.}

\begin{equation} \label{Eq:Rmsprop1}
\hspace{0.1cm}E(g^2)_{t} = \beta E(g^2)_{t-1} + (1-\beta)(\frac{\partial C}{\partial w})^2 \hspace{2.4cm}
\end{equation}

\hspace{-0.5cm}where, $E[g]$ is moving average of squared gradients,  $\frac{\partial C}{\partial w}$ is gradient of the cost function with respect to the weight, $\eta$ learning rate, $\beta$ is moving average parameter.

\begin{equation} \label{Eq:Rmsprop2}
w_{t} = w_{t-1} - \frac{\eta}{\sqrt{E(g^2)_{t}}}\frac{\partial C}{\partial w}
\end{equation}
$w_{t}$ is value of the weight parameter at iteration t.

\vspace{0.5cm}

\begin{table}
\caption{Parameters of learning algorithm
used in MSCNN based architecture.} \label{tab:Parameters}
\vspace{0.3cm}
\centering
\resizebox{\columnwidth}{!}{%
\begin{tabular}{c c}
\hline 
Parameter name & Parameter value \\
\hline
 Learning algorithm & RMSProp \\ [0.2cm]
 Initial learning rate & 0.001 \\ [0.2cm]
 Learning decay rate & 0.993 \\ [0.2cm]
 Dropout & 0.5 \\ [0.2cm]
 Batch size & 500 \\ [0.2cm]
 Total number of training epochs & 500 \\
\hline
\end{tabular}%
}
\end{table}

{An adaptive learning process RMSProp \cite{ruder2016overview,basu2018convergence} learning algorithm is used to train our proposed system.} All the three columns are trained simultaneously against same cost calculated at the final FC layer. {For calculating the loss during training crossentropyLoss \cite{farahnak2016multi} is used as the loss function (shown in \ref{Eq:CEL}).} The proposed architecture is trained for 500 epochs with variable learning rate \cite{cirecsan2012multi}, in which the learning rate is decreased by a factor of 0.993/epoch until it reaches to the value of 0.00003. As suggested by LeCun et al. \cite{lecun2012efficient}, the dataset is randomly shuffled before each epoch of RMSProp based training. Shuffling introduces heterogeneity in the datasets and increases the convergence rate of the learning algorithm. The proposed architecture is trained until the error rate converges or the total number of epochs reaches the aforementioned maximum. Dropout regularization \cite{srivastava2014dropout,wager2013dropout} is used (only in the FC layers except the final FC layer) to train the proposed architecture to help reduce the chances of overfitting of the network while training. The parameter of the learning algorithm used to train the proposed architecture is presented in \ref{tab:Parameters}. After the training is over, an SVM based classifier with Gaussian kernel is used to perform the classification tasks. SVM based classifier takes the output feature vector from the final FC layer of the proposed architecture as input and after completion of training, it predicts the actual class label of each test data.
\vspace{0.5cm}

\begin{equation} \label{Eq:CEL}
H(y,\hat{y}) = -\sum_{i=1}^{k}[y_{i}\ln{\hat{y_{i}}}]
\end{equation}
where, $y$ is actual class, $\hat{y}$ is predicted class, $k$ is total number of classes.
\vspace{0.5cm}

As we create a validation set randomly from the training images for every dataset used in our experiments, the size of the training dataset is reduced and consequently the final training set becomes insufficient to train the whole architecture. To overcome this problem, we first divide the initial training set in train set and validation set as done before. The proposed architecture is now trained on training set and saved against the best accuracy achieved on validation set. Now the epoch number is noted for the best validation accuracy. Now the network is again trained on the initial training set (before creating the validation set) until the noted epoch number is achieved and the network is saved. Now the architecture is test against the test set as done previously.

\begin{table*}
\caption{Datasets used in the present work.} \label{tab:Dataset}
\vspace{0.1cm}
\centering
\resizebox{\linewidth}{!}{%
\begin{tabular}{c c c c c c }
\hline 
Index & Name of the dataset & Dataset type & Number of training samples & Number of test samples & Reference \\
\hline
 D1 & CMATERdb 3.1.1 & Bangla digits & 4000 & 2000 & \unskip~\cite{das2012statistical} \\ [0.2cm]
 
 D2 & ISIBanglaDigit & ISI numerals & 23,500 & 4000 & \unskip~\cite{bhattacharya2009handwritten} \\ [0.2cm]

 D3 & CMATERdb 3.1.2 & Bangla basic characters & 12,000 & 3000 & \unskip~\cite{sarkar2012cmaterdb1} \\ [0.2cm]

 D4 & CMATERdb 3.1.3 & Bangla compound characters & 34,229 & 8468 & \unskip~\cite{das2014benchmark} \\

\hline
\end{tabular}%
}
\end{table*}

\subsection{Extraction of multi-scale feature}

Once the network is trained and connection weights are learnt, the test image is propagated through each of the column of the architecture separately. As the columns are independent from each other, they extract features independently in the corresponding convolutional and pooling layers. {As mentioned earlier, the feature maps sampled at every level of different columns are combined and more useful features are extracted through an FC layers (as in \ref{fig:combination}). After performing feature extraction through every column, the features are combined to produce the resultant feature vector which is then forward propagated through a final FC layer to extract final feature descriptors. A softmax classifier is trained with the finally extracted feature (as in \ref{fig:Architecture}) and after training, the classifier is used to predict the final class label of the test image. Hence, the final concatenated global and local features is used as the final feature descriptors of the original input images in this present architecture.} As mentioned above, a multi-class SVM classifier is used for classification instead of an MLP based classifier because SVM performs better than MLP in handwritten digits or characters recognition tasks \cite{das2010handwritten}.

Every convolutional and maxpooling layer of the proposed architecture decreases the feature map size from its previous layer. Generally, most of the information loss occurs due to the maxpooling layers. For this reason, it is imperative to used smaller kernels (i.e. $2 \times 2$) in the pooling layers. As the convolutional layers are less problematic than maxpooling layers, the larger kernel (i.e. $3 \times 3$, $5 \times 5$, $7 \times 7$) in the convolutional layers are admissible.

In order to extract both global and local features from multiple scales different kernel sizes are used at convolutional layers of every level of different columns (as in \ref{fig:Multi_Column}). This allows multi-scaled feature extraction within same level. Also, for a definite column the consecutive convolutional layers have kernel sizes different from each other, which allows multi-scaled feature extraction within each column \cite{sarkhel2017multi}. The combination of these multi-scale features represents better abstraction of the input image than the features from fixed scales. These multi-scaled global and local features extracted through two different types (i.e. \textit{level-wise multi-scaling} and \textit{column-wise multi-scaling}) of multi-scaling approach are used to recognize different stylistically varying Bangla handwritten characters and digits.\vspace{-0.2cm}

\section{Experiments}\label{sec:Experimental Results}
As mentioned before, a non-explicit global and local feature extraction technique has been proposed in this present work. For this purpose a MSCNN based architecture is used for recognition of handwritten characters and digits from multiple Bangla scripts. A Python based library, PyTorch is used to implement and train the proposed MSCNN based architecture. Basic image processing operations are performed using MATLAB. All of the experiments are performed using systems with Intel dual core i5 processors, 8 GB RAM and a NVIDIA GeForce 1050 Ti graphics card with 12 GB internal memory.

\subsection{Datasets}The proposed architecture is tested on four publicly available datasets on isolated handwritten characters and digits, belonging to popular Bangla scripts. Details of the datasets used in our current experiment is shown in the \ref{tab:Dataset}. These significantly intricate datasets, which are different from each other belonging to Bangla scripts on isolated handwritten characters or digits comprise an ideal test suite for the present work. More details of these datasets can be found in references cited in the right most column of the table.

\begin{table*}
\caption{Recognition accuracy achieved by our proposed architecture on different datasets.} \label{tab:Result_PA}
\centering
\resizebox{\linewidth}{!}{%
\begin{tabular}{c c c c c c}
\hline
Name of the dataset & Dataset type & Number of classes & Random validation set (\%) & $5$-Fold cross-validation (\%) & $10$-Fold cross-validation (\%) \\
\hline
 CMATERdb 3.1.1 & Bangla digits & 10 & 98.15 \% & 98.18 \% & \textbf{98.27 \%} \\ [0.2cm]
 
 ISIBanglaDigit & ISI numerals & 10 & 99.36 \% & 99.34 \% & \textbf{99.38 \%} \\ [0.2cm]

 CMATERdb 3.1.2 & Bangla basic characters & 50 & 96.65 \% & 96.56 \% & \textbf{96.7 \%} \\ [0.2cm]

 CMATERdb 3.1.3 & Bangla compound characters & 171 & 93.48 \% & 93.5 \% & \textbf{93.53 \%} \\

\hline
\end{tabular}%
}
\end{table*}\vspace{-0.25cm}

\subsection{Pre-processing of datasets}Images of each datasets are passed through a few numbers of pre-processing steps. Every image on isolated handwritten characters or digits is binarized and centre-cropped by the tightest bounding box \cite{sarkhel2017multi} and finally resized to $32 \times 32$ pixels and normalized with mean 0.5 and std 0.5. Also, noise in the images are removed by combination of median and Gaussian filter. 

After the pre-processing step, the training dataset is randomly divided into training set and validation set in such a way that the size of the validation set matches with the size of the test set. Now, the architecture is trained on the training set and saved against the best recognition accuracy achieved on validation set. After the network is trained, it is evaluated on the test set.

\subsection{Experimental results}
The experimental results on every dataset is presented in this section. As mentioned earlier, the proposed methodology is tested on four publicly available benchmark datasets on isolated handwritten characters or digits on Bangla scripts. Among these four datasets two are belonging to Bangla digits, one is on Bangla basic characters and last one is on Bangla compound characters. Details of these datasets are given in \ref{tab:Dataset}. {The proposed MSCNN based architecture is applied on each of these four datasets 15 times and the SVM based classifier performed the classification task every time based on the extracted features from CNN based architecture. We have also presented recognition accuracies achieved by our proposed architecture using \textit{5-Fold and 10-Fold cross-validation} techniques on the training set. And the best classification result achieved among all the trials for every dataset is listed in \ref{tab:Result_PA}.} From our experimental observation we have found that the results achieved by the present architecture is better than the results achieved by the architecture which uses only the local features of the input images. For comparison the results achieved by using only local features are given in \ref{tab:Result_LA}.\vspace{-0.2cm}

\subsection{Comparative analysis}\label{sec:Comparative Analysis}
In this section we described various types of comparisons with our proposed architecture. As, mentioned before the proposed network uses a feature fusion technique to combine global and local features sampled at each convolutional layer of the network. There can be other kinds of feature combination methods other than our proposed feature fusion technique. We presented some of those combination techniques in \ref{sec:Different combintion} and did a comparison based study to prove the superiority of our proposed combination technique. Also, to prove the efficacy of using global and local features for recognition handwritten characters or digits, we compared our proposed architecture with some of most popular contemporary works in \ref{sec:contemporary works}.

\begin{table*}
\caption{Comparison between recognition accuracies achieved by the architecture which use only local features and the proposed architecture.} \label{tab:Result_LA}
\vspace{0.05cm}
\centering
\resizebox{\linewidth}{!}{%
\begin{tabular}{c c c c}
\hline
Name of the dataset & Dataset type & Proposed architecture (\%) & Using local features only (\%) \\
\hline
 CMATERdb 3.1.1 & Bangla digits & \textbf{98.15 \%} & 95.75 \% \\ [0.2cm]
 
 ISIBanglaDigit & ISI numerals & \textbf{99.36 \%} & 98.68 \% \\ [0.2cm]

 CMATERdb 3.1.2 & Bangla basic characters & \textbf{96.65 \%} & 94.45 \% \\ [0.2cm]

 CMATERdb 3.1.3 & Bangla compound characters & \textbf{93.48 \%} & 90.56 \% \\

\hline
\end{tabular}%
}
\end{table*}

\begin{table*}
\vspace{0.2cm}
\caption{Comparison between the recognition accuracies achieved by the proposed architecture and different feature combination techniques.} \label{tab:Result_All}
\vspace{0.05cm}
\centering
\resizebox{\linewidth}{!}{%
\begin{tabular}{c c c c c c}
\hline 
Name of the dataset & Dataset type & First baseline (\%) & Second baseline (\%) & Third baseline (\%) &  Proposed architecture (\%) \\
\hline
 CMATERdb 3.1.1 & Bangla digits & 97.15 \% & 97.76 \% & 97.52 \% & \textbf{98.15 \%} \\ [0.2cm]
 
 ISIBanglaDigit & ISI numerals & 99.12 \% & 99.18 \% & 99.14 \% & \textbf{99.36 \%} \\ [0.2cm]

 CMATERdb 3.1.2 & Bangla basic characters & 95.92 \% & 96.43 \% & 95.96\% & \textbf{96.65 \%} \\ [0.2cm]

 CMATERdb 3.1.3 & Bangla compound characters & 91.15 \% & 92.34 \% & 91.46 \% & \textbf{93.48 \%} \\

\hline
\end{tabular}%
}
\end{table*}

\begin{small}
\begin{table*}
\caption{A comparative analysis of the proposed methodology with some of the popular contemporaries.} \label{tab:Result_Contemp}
\vspace{0.3cm}
\centering
\resizebox{\linewidth}{!}{%
\begin{tabular}{c c c}
\hline
Dataset type & Work reference & Recognition accuracy (\%) \\
\hline
 Bangla digits & Das et al. \unskip~\cite{das2012genetic} & 97.80 \% \\
               & Basu et al. \unskip~\cite{basu2012mlp} & 96.67 \% \\
               & Roy et al. \unskip~\cite{roy2012new} & 95.08 \% \\
               & Roy et al. \unskip~\cite{roy2014axiomatic} & 97.45 \% \\
               & Lecun et al. \unskip~\cite{le1994word} & 97.31 \% \\
               & The Present Work & \textbf{98.15} \% \\
\hline
ISI numerals & Sharif et al. \unskip~\cite{sharif2016hybrid} & 99.02 \% \\
                & Wen et al. \unskip~\cite{wen2012classifier} & 96.91 \% \\
                & Das et al. \unskip~\cite{das2012genetic} & 97.70 \% \\
                & Akhnad et al. \unskip~\cite{rahman2015bangla} & 97.93 \% \\
                & CNNAP \unskip~\cite{akhand2016convolutional} & 98.98 \% \\
                & Roy et al. \unskip~\cite{roy2012new} & 93.338 \% \\
                & Rajashekararadhya  et al. \unskip~\cite{rajashekararadhya2009zone} & 94.20 \% \\
                & The Present Work & \textbf{99.36} \% \\

\hline
Bangla basic characters & Roy et al. \unskip~\cite{roy2012region} & 86.40 \% \\
                        & Das et al. \unskip~\cite{das2010handwritten} & 80.50 \% \\
                        & Basu et al. \unskip~\cite{basu2009hierarchical} & 80.58 \% \\
                        & Sarkhel et al. \unskip~\cite{sarkhel2015enhanced} & 86.53 \% \\
                        & Bhattacharya et al. \unskip~\cite{bhattacharya2006recognition} & 92.15 \% \\
                        & Lecun et al. \unskip~\cite{le1994word} & 92.88 \% \\
                        & The Present Work & \textbf{96.65} \% \\
\hline
Bangla compound characters & Das et al. \unskip~\cite{das2010handwritten} & 75.05 \% \\
                           & Das et al. \unskip~\cite{das2015handwritten} & 87.50 \% \\
                           & Sarkhel et al. \unskip~\cite{sarkhel2015enhanced} & 78.38 \% \\
                           & Sarkhel et al. \unskip~\cite{sarkhel2016multi} & 86.64 \% \\
                           & Pal et al. \unskip~\cite{pal2015recognition} & 93.12 \% \\
                           & Lecun et al. \unskip~\cite{le1994word} & 86.85 \% \\
                           & Roy et al . \unskip~\cite{roy2017handwritten} & 90.33 \% \\
                           & The Present Work & \textbf{93.48} \% \\
\hline
\end{tabular}%
}
\end{table*}
\end{small}

\subsubsection{Comparing Different Column Combination Techniques}\label{sec:Different combintion}
The proposed system has been performed on five datasets and significantly improved results have been achieved. In the proposed multi-column CNN based architecture various combination techniques of combining global and local features are possible. Some of the combination techniques are presented in this section and a comparison based study on the recognition accuracy achieved by using these techniques is taken up.

\paragraph{\textbf{First baseline:} }

The proposed multi-column CNN architecture has three columns as in \ref{fig:combination} and from every convolutional layer features are extracted through FC layers. These extracted features are combined to generate the final feature descriptor which is then used for recognition of multiple handwritten characters as mentioned in \ref{sec:Present Work}. Other than the proposed level-wise feature combination technique (shown in \ref{fig:Architecture}) the features can be combined by concatenating all the feature vectors sampled at every local FC layer and finally forward propagated through a final FC layer with 2048 output neurons to generate the final output feature vector.
The recognition accuracy achieved by the first baseline architecture on all the five datasets are listed in \ref{tab:Result_All}.

\paragraph{\textbf{Second baseline:} }Feature vectors extracted at all the local FC layers of the proposed multi-column architecture (as shown in \ref{fig:Architecture}) can also be combined with a column-wise feature combination technique, i.e. features from every local FC layer of a definite column are concatenated. Features sampled at every local FC layers of first column are concatenated and passed through 2048FC (FC layer with 2048 output neurons). Similarly, concatenated features from second and third column are forward propagated through 5120FC and 8192FC respectively. Finally, the features from these columns are combined with the input features extracted by a 512FC layer and the final concatenated feature is passed through a final 2048FC layer. The recognition accuracy achieved by the second baseline architecture on all the five datasets are given in \ref{tab:Result_All}.\vspace{-0.2cm}

\paragraph{\textbf{Third baseline:} }As mentioned earlier the proposed architecture consists of three columns (as in \ref{fig:combination}) and features from every local FC layer are concatenated to generate final feature descriptor. In the third combination technique the feature maps from convolutional layers are combined with feature-concatenation technique (shown in \ref{fig:concatenation}). An extra skip connection is introduced between the convolutional layers of each column of our proposed architecture. The output feature maps of the first level is combined with the output feature maps of the second level for every column to form the input feature maps to the third level.
The recognition accuracy achieved by the third baseline architecture is given in \ref{tab:Result_All}. The graphical abstracts of the architectures described in the above baselines are illustrated in details in the accompanying supplementary files.

\begin{table*}
\vspace{0.7cm}
\caption{Best performance on the test set using different data augmentation techniques} \label{tab:Result_Data_Aug}
\vspace{0.3cm}
\centering
\resizebox{\linewidth}{!}{%
\begin{tabular}{c c c c c }
\hline 
Transformation & Bangla digits & ISI numerals & Bangla basic characters & Bangla compound characters \\
\hline
None & 98.15 \% & 99.36 \% & 96.65 \% & 93.48 \% \\ [0.2cm]

ColourJitter & \textbf{98.70} \% & 99.38 \% & \textbf{96.85} \% & 93.12 \% \\ [0.2cm]

Random Horizontal Flip & 98.10 \% & 99.32 \% & 96.70 \% & 92.88 \% \\ [0.2cm]

Random Vertical Flip & 98.54 \% & 99.22 \% & 96.20 \% & 92.82 \% \\ [0.2cm]

Random Crop & 98.60 \% & \textbf{99.40} \% & 95.20 \% & 93.55 \% \\ [0.2cm]

Rotation & 98.52 \% & 99.34 \% & 96.63 \% & \textbf{94.30} \% \\ [0.2cm]

Random Affine Transformation & 98.20 \% & 99.17 \% & 94.45 \% & 92.12 \% \\
\hline
 
\end{tabular}%
}
\end{table*}

\begin{table}
\vspace{0.3cm}
\caption{Comparison with recognition accuracy between different training methods.} \label{tab:Result_SVS}
\vspace{0.1cm}
\centering
\resizebox{\linewidth}{!}{%
\begin{tabular}{c c c}
\hline 
Dataset type & Simultaneous Training & Separate Training \\
\hline
Bangla digit & \textbf{98.15 \%} & 97.55 \% \\ [0.2cm]

ISI numerals & \textbf{99.36 \%} & 98.25 \% \\ [0.2cm]

Bangla basic character & \textbf{96.65 \%} & 95.88 \% \\ [0.2cm]

Bangla compound character & \textbf{93.48 \%} & 91.75 \% \\
\hline
\end{tabular}%
}
\end{table}

\subsubsection{Comparisons against contemporary works}\label{sec:contemporary works}
Significantly improved results have been achieved by the proposed system for all of the datasets used in our experimental setup. To prove the superiority of the proposed method, its performance is compared with some of the popular, contemporary works. The best recognition accuracy achieved by a contemporary system is shown in boldface. The comparative analysis is given in \ref{tab:Result_Contemp}.\vspace{-0.2cm}

\subsection{Effects of different pre-processing and training techniques}
In this section we presented some of popular data augmentation techniques to improve the performance of the proposed architecture and showed a comparison between the recognition accuracy achieved with different data augmentation techniques. This section also presented different training techniques to train a multi-column CNN based architecture.

\subsubsection{Data augmentation}
Sometimes to improve the performance each input is stochastically transformed during training a CNN-based architecture [9, 12]. We used 6 different types of transformations (e.g. crop, rotation, flip etc) in our current experimental setup. Each of the transformation is applied separately to train our proposed architecture and the recognition accuracy is presented in \ref{tab:Result_Data_Aug}. The best recognition accuracy on each dataset is shown in boldface.

With \textit{colourjitter} we randomly change the brightness, contrast and saturation of input images with a factor of 0.05. With \textit{random horizontal flip} we randomly flip input images horizontally with probability 0.5. Similarly, with \textit{random vertical flip} we randomly flip input images vertically with probability 0.5. With \textit{random crop} we crop the images at random locations. With \textit{rotation} the training images were rotated by an angle of \ang{20}. With \textit{random affine transformation} we did affine transformation randomly between the range of [\ang{-45}, \ang{45}] of each image with the center invariant.\vspace{-0.15cm}

\subsubsection{Training methods}
As mentioned earlier, our proposed architecture is trained with an end-to-end training method, in which all the three columns are simultaneously trained against same loss calculated at the final FC layer of the architecture. {Another common practice to train a multi-column CNN based architecture is to train each column separately and finally merge all the columns to predict the final class label of each test data. In the later training technique, first every column of our proposed architecture is trained separately on train set and saved against the best recognition accuracy achieved on validation set. After all the three columns are trained, the local FC layers of all convolutional layers along with other FC layers are trained as follows : batches of training data are forward propagated through every previously trained column and the features maps generated at each sampling layer is passed through its local FC layer and these feature maps are gradually extracted through the other FC layers. The connection weights between the FC layers are updated using backpropagation against the loss calculated at final FC layer of our proposed architecture. Finally, the connection weights between all the FC layers are saved against the minimum loss achieved on validation set. We tested the performances of both the training techniques and found that our proposed network performs better in case of simultaneous training strategy.} For comparison, the performance of both training methods on the test set are shown in \ref{tab:Result_SVS}.

\section{Conclusion}
As mentioned in our present work, a multi-scale multi-column skip convolutional neural network based architecture is proposed for recognition of various handwritten characters and digits. This architecture uses a combination of multi-scale global and local geometric features of pattern images for generating ubiquitous pattern which describes the pattern images in more precise way. After the exploratory experiments on different methods on multi-column CNN based architecture, we conclude that the multi-scale global and local features of a pattern image together can represent more robust and impeccable description of the original image. Significantly better results on recognition accuracy advocate the effectiveness of the proposed methodology. This proposed methodology opens a new area in research towards pattern recognition tasks on different types of pattern images.

\section*{Acknowledgments}
The authors are thankful to the Center for Microprocessor Application for Training Education and Research (CMATER) and Project on Storage Retrieval and Understanding of Video for Multi- media (SRUVM) of Computer Science and Engineering Department, Jadavpur University, for providing infrastructure facilities during progress of the work. The current work, reported here, has been partially funded by University with Potential for Excellence (UPE), Phase-II, UGC, Government of India.

\section*{Conflict of interest declaration}
The authors certify that they have no affiliations with or involvement in any organization or entity with any financial interest (such as honoraria; participation in speakers’ bureaus; membership, consultancies, stock ownership, or other equity interest; and expert testimony or patent-licensing arrangements), or non-financial interest (such as personal or professional relationships, affiliations, knowledge or beliefs) in the subject matter or materials discussed in this manuscript to the best of their knowledge.


\balance

\bibliographystyle{spphys}
\bibliography{References}

\end{document}